\documentclass[runningheads]{llncs}
\usepackage{graphicx}
\usepackage{times}

\usepackage{soul}
\usepackage{url}
\usepackage[hidelinks]{hyperref}
\usepackage[utf8]{inputenc}
\usepackage[small]{caption}
\usepackage{graphicx}
\usepackage{amsmath}
\usepackage{booktabs}
\usepackage[utf8]{inputenc} 
\usepackage[T1]{fontenc}    
\usepackage{booktabs}       
\usepackage{amsfonts}       
\usepackage{nicefrac}       
\usepackage{microtype}      
\usepackage{lipsum}
\usepackage{multirow}
\usepackage{siunitx}

\begin{document}
\title{IPOD: An Industrial and Professional Occupations Dataset and its Applications to Occupational Data Mining and Analysis}
%

%
%

\author{
Junhua Liu\inst{1} \and 
Yung Chuen Ng\inst{2} \and 
Kristin L. Wood\inst{1} \and 
Kwan Hui Lim\inst{1}
}


\authorrunning{J. Liu et al.}

%

\institute{
 Singapore University of Technology and Design\\
 \email{junhua\_liu@mymail.sutd.edu.sg, kristinwood@sutd.edu.sg, kwanhui\_lim@sutd.edu.sg} \and
 National University of Singapore\\
 \email{e0201912@u.nus.edu}}
 
%
\maketitle              
\begin{abstract}
Occupational data mining and analysis is an important task in understanding today's industry and job market. Various machine learning techniques are proposed and gradually deployed to improve companies' operations for upstream tasks, such as employee churn prediction, career trajectory modelling and automated interview. Job titles analysis and embedding, as the fundamental building blocks, are crucial upstream tasks to address these occupational data mining and analysis problems. In this work, we present the Industrial and Professional Occupations Dataset (IPOD), which consists of over 190,000 job titles crawled from over 56,000 profiles from Linkedin. We also illustrate the usefulness of IPOD by addressing two challenging upstream tasks, including: (i) proposing \textit{Title2vec}, a contextual job title vector representation using a bidirectional Language Model (biLM) approach; and (ii) addressing the important occupational Named Entity Recognition problem using Conditional Random Fields (CRF) and bidirectional Long Short-Term Memory with CRF (LSTM-CRF). Both CRF and LSTM-CRF outperform human and baselines in both exact-match accuracy and F1 scores. The dataset and pre-trained embeddings are available at https://www.github.com/junhua/ipod.

\end{abstract}
\begin{table}[h]
\centering
\begin{tabular}{@{}llll@{}}
\toprule
Literature & Source & Size & Avail. \\ \midrule
\textbf{\begin{tabular}[c]{@{}l@{}}IPOD \end{tabular}} & \textbf{Linkedin} & \textbf{190K} & \textbf{Yes} \\ \midrule
Mimno et al., 2008 & Resumes & 54K & No \\
Lou et al., 2010 & Linkedin & 67K & No \\
Paparrizos et al., 2011 & Web & 5M & No \\
Zhang et al., 2014 & Job site & 7K & No \\
Liu et al., 2016 & Social network & 30K & No \\
Li et al., 2017 & Linkedin & - & No \\
Li et al., 2017 & High tech co. & - & No \\
Yang et al., 2017 & Resumes & 823K & No \\
zhu et al., 2018 & Job portals & 2M & No \\
James et al., 2018 & APS & 60K & Yes \\
Yang et al., 2018 & Var. channels & - & No \\
Xu et al., 2018 & Pro. networks & 20M & No \\
Qin et al., 2018 & High tech co. & 1M & No \\
Lim et al., 2018 & Linkedin & 10K & No \\
Shen et al., 2018 & High tech co. & 14K & No \\ \bottomrule
\end{tabular}
\caption{A survey of datasets used for related works. No available datasets can be found publicly except a dataset of publications and authors from American Physics Society (APS)~\cite{james2018prediction} that only describes the names and affiliations of physics scientists without titles.}
\label{tab:dataset}
\end{table}

\section{Introduction}
\label{sec:introduction}

There is a growing interest in occupational data mining and analysis tasks in recent years, especially with the rapid digitization of today's economy and jobs. Furthermore, the advancement of AI and robotics are changing every industry and every sector, challenging the employability of work force especially those with high level of repetition. 

Occupational data mining and analysis is also an important topic in academia research. In the literature, various downstream tasks of occupational analysis showed promising results using machine learning techniques, such as employee churn prediction ~\cite{james2018prediction,yang2018one,zhao2018employee}, professional career trajectory modelling~\cite{liu2016fortune,mimno2008modeling} and predicting employee behaviors with various factors~\cite{chen2012mining,cetintas2011identifying}, among others. 

These earlier works on occupational data mining and analysis fulfil industrial demands and create substantial value for both the companies and professionals. However, these tasks remain challenging due to a lack of publicly available datasets. For many years, the relevant data resides with a small number of enterprises, which utilize such data privately for maintaining their competitive edge in the industry. Thus, such data is not publicly available for smaller companies or individuals to better understand the industry and job market or for career planning purposes. Table~\ref{tab:dataset} shows a survey of 15 related works that utilizes similar types of dataset, of which only one is publicly available (apart from our proposed dataset). 

\subsection{Main Contributions}

In this paper, we make the following contributions:
\begin{itemize}
    \item To address the needs of career analysis for industry, we present and make publicly available the Industrial and Professional Occupation Dataset (IPOD). This dataset consists of 192,295 occupation entries, drafted by working professionals for their Linkedin profiles, with the motivations of displaying their career achievement, attracting recruiters or expanding professional networks. As shown in Table~\ref{tab:dataset}, IPOD is the largest publicly available dataset out of a total of 16 used in various recent works.
    \item To improve the usability of IPOD, we propose \textit{Title2vec}, a contextual job title vector representation with a bidirectional Language Model (biLM)~\cite{peters2018deep} approach. This upstream embedding task map the raw job titles into a high-dimensional vector space that allows and boosts the performance of the downstream occupational NER task.
    \item To further demonstrate the usefulness of IPOD, we propose two models for a challenging Named Entity Recognition (NER) task, alongside with 2 baselines and human performance. The two models include a probabilistic machine learning model, namely Conditional Random Field (CRF), and a state-of-the-art recurrent neural network model, namely bidirectional LSTM-CRF~\cite{liu2018empower}. Both of two models outperform human and baselines in terms of Exact Match (EM) accuracy and F1 scores for both overall and tag-specific results.
\end{itemize}

Existing corpora for Named Entity Recognition (NER) tasks~\cite{finkel2005incorporating,sang2003introduction,weischedel2013ontonotes,borchmann2018approaching} typically use general tags such as \textbf{LOC}ation, \textbf{PER}son, \textbf{ORG}anization, \textbf{MISC}ellaneous, etc.. On the contrary, IPOD provides domain-specific NE tags to denote the properties of occupations, such as \textit{\textbf{RES}ponsibility}, \textit{\textbf{FUN}ction} and \textit{\textbf{LOC}ation}. All named entities are tagged using a comprehensive gazetteer created by three experts, which reports high inter-rater reliability, achieving 0.853 on Percentage Agreement~\cite{viera2005understanding} and 0.778 on Cohen’s Kappa~\cite{artstein2008inter}, with no instances where all three annotators disagree. The labels are further processed by adding prefix using BIOES tagging scheme~\cite{Ratinov2009design}, i.e., \textbf{B}egin, \textbf{I}nside, \textbf{E}nding, \textbf{S}ingle, and \textbf{O} indicating that a token belongs to no chunk, indicating the positional features of each token in a title.

\section{Description of the Industrial and Professional Occupations Dataset (IPOD)}
\label{sec:collection}

In this section, we describe our data collection process, characteristics of the IPOD dataset and results from an exploratory data analysis.

\subsection{Data Collection}
We obtained over 192K job titles based on Linkedin profiles from Asia and the United States, as representatives of the world's most competitive economies~\cite{akhtar_2019}.  Subsequently, the raw data underwent a series of processing, including converting to lowercase, substituting meaningful punctuation to words (i.e. changing $\&$ to $and$) and removing special symbols. We decided not to lemmatize or stem the words because the original forms suggest its most accurate named entity, i.e., strategist is labeled as RES while strategy is labeled as FUN.

\subsection{Dataset Analysis}

\begin{figure*}[t]
\centering
  \includegraphics[width=\textwidth]{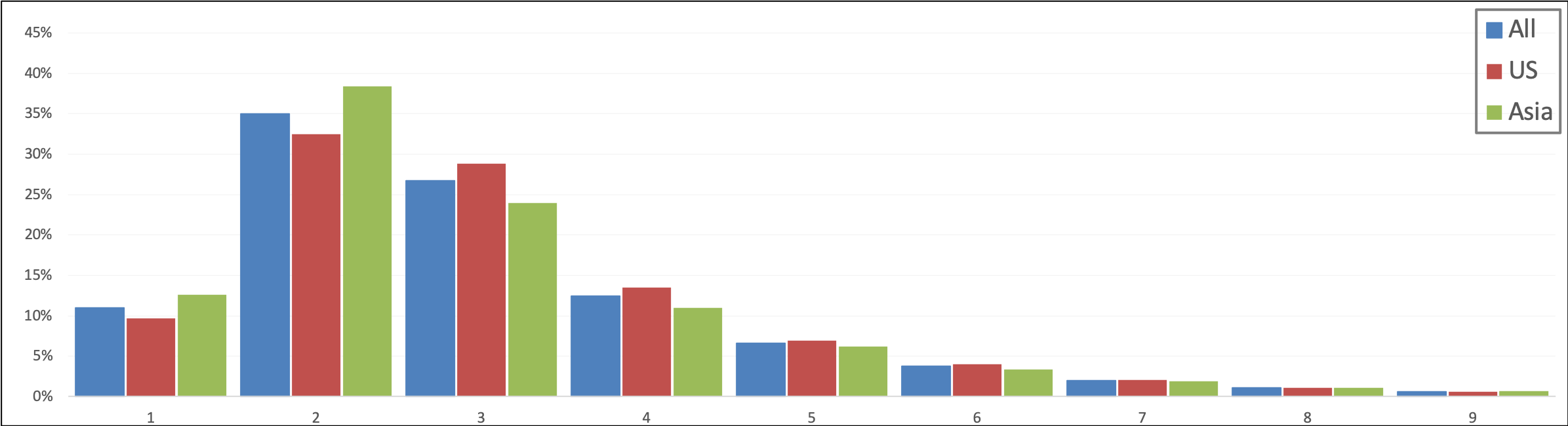}
  \caption{Histogram of occupation entries. The x-axis shows the number of words in the occupation title and the y-axis shows the usage frequency in percentage.}
  \label{fig:lhist}
\end{figure*}

This section discusses the exploratory data analysis conducted to better understand the properties of IPOD. The statistics and histogram of the length of job titles can be found in Table~\ref{tab:lstats} and Fig.~\ref{fig:lhist} respectively. The corpus comprises of 192,295 English occupation entries from 56,648 unique profiles. These profiles are mainly from United States ($56.7\%$) and Asia ($43.3\%$). Most of the titles fall within five words, contributing to 91.7\% of the entries, as shown in both Table~\ref{tab:lstats} and Fig.~\ref{fig:lhist}). The median statistics and the histogram also suggest that job titles written by Asian professionals tend to be shorter, i.e., within two words, than that by US professionals. 

\begin{table}[h]
\parbox{.5\linewidth}{
\centering
\begin{tabular}{@{}cccc@{}}
\toprule
 & All & US & Asia \\ \midrule
min & 1 & 1 & 1 \\
max & 21 & 17 & 21  \\
avg & 3.0 & 3.1 & 2.9  \\
med & 3 & 3 & 2 \\ \bottomrule
\end{tabular}
\caption{Statistics of entries}
\label{tab:lstats}
}
\hfill
\parbox{.5\linewidth}{
\centering
\begin{tabular}{@{}cc@{}}
\toprule
NE & Count \\ \midrule
RES & 310570 \\
FUN & 255974 \\
LOC & 9998 \\
O & 66948 \\ \bottomrule
\end{tabular}%
\caption{NE counts}
\label{tab:labels}
}
\end{table}

\begin{figure*}[ht]
  \includegraphics[width=\textwidth]{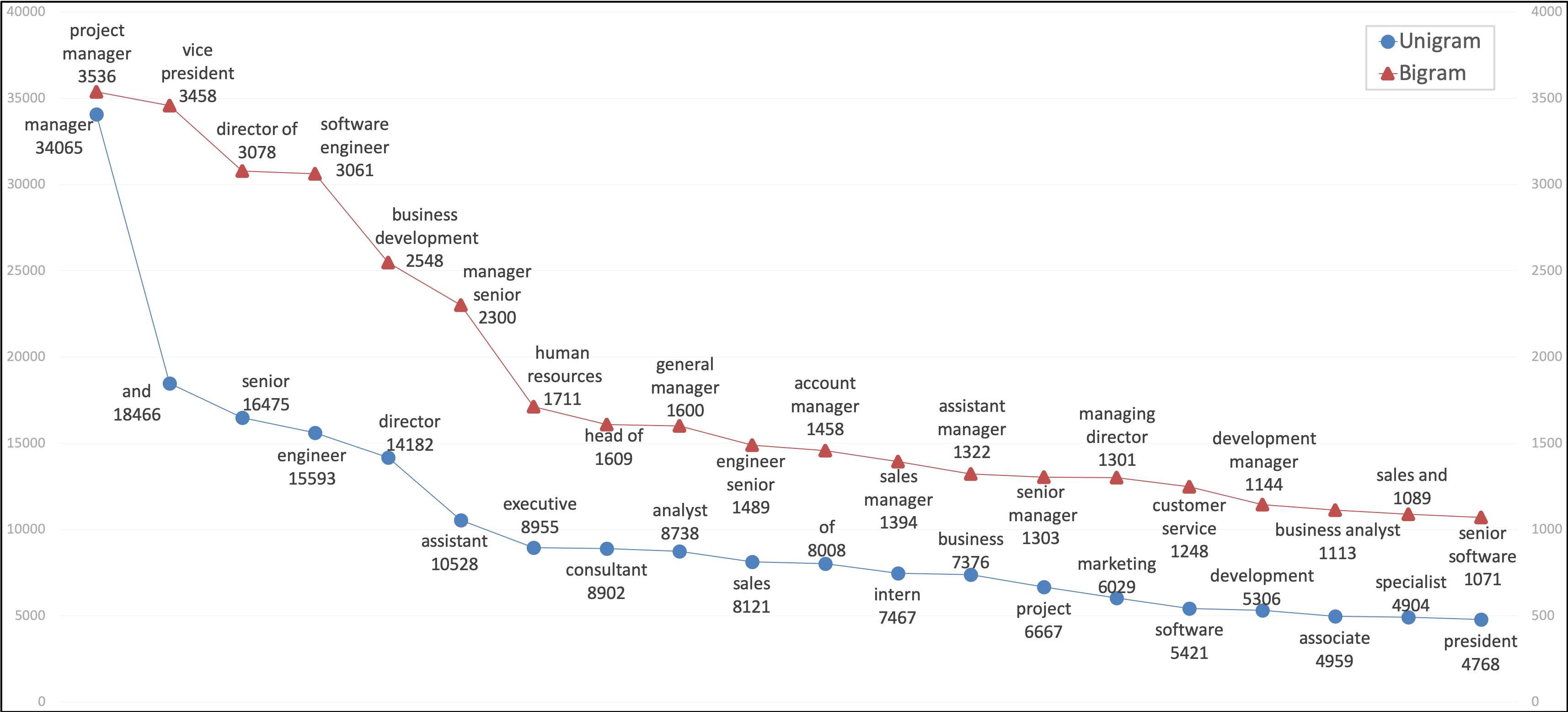}
  \caption{N-grams analysis of the occupation title entries, where the y-axis shows the usage frequency.}
  \label{fig:ngrams}
\end{figure*}

Figure~\ref{fig:ngrams} shows the distribution of top 20 Unigrams and Bigrams~\cite{damashek1995gauging} of IPOD. In the Unigram case, the most popular token, \textit{manager}, appears in 34,065 entries, about twice as much as the next few popular ones, i.e., \textit{and (18,466)}, \textit{senior (16,475)}, \textit{engineer (15,593)} and \textit{director (14,182)}. On the contrary, the Bigram case shows a gentler curve (i.e. lower slope), with \textit{project manager (3,536)} and \textit{vice president (3,458)} being the top two choices.

\subsection{Domain-specific Sequence Tagging}

Job titles serve as a concise indicator for one's level of responsibility, seniority and scope of work, described with a combination of \textit{responsibility}, \textit{function} and \textit{location}. Table~\ref{tab:NEsample} shows examples of the occupational NE tags.

\begin{figure}[t]
  \includegraphics[width=0.8\textwidth]{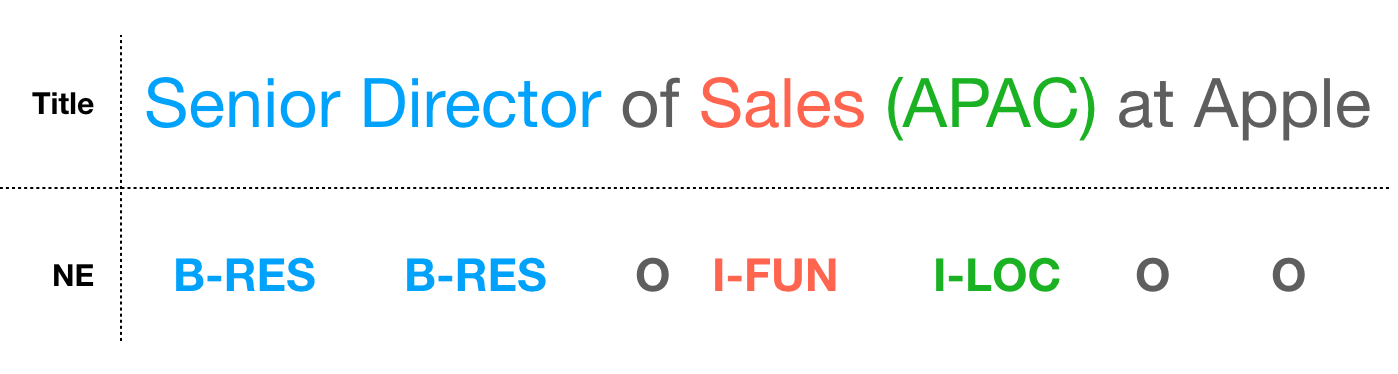}
  \centering
  \caption{An example of occupational title and its domain-specific NE tags. Tokens in a title indicate the person's responsibility (\textbf{RES}), function (\textbf{FUN}), and location (\textbf{LOC}). Furthermore, the NE tags are also added with positional prefixes using BIOES scheme, i.e., \textbf{B}egin, \textbf{I}nside, \textbf{O}thers, \textbf{E}nding and \textbf{S}ingle.}
  \label{fig:NE}
\end{figure}
 
\textbf{Responsibility}, as its name suggests, describes the role and duty of a working professional. As shown in figure~\ref{fig:NE}, responsibility may include indicators of managerial levels, such as \textit{director, manager} and \textit{lead}, seniority levels, such as \textit{vice, assistant} and \textit{associate}, and operational role, such as \textit{engineer, accountant} and \textit{technician}. A combination of the three sub-categories draws the full picture of one's responsibility.

\textbf{Function} describes business functions in various dimensions. Specifically, \textit{Departments} describes the company's departments the staffers are in, such as \textit{sales, marketing} and \textit{operations}; \textit{Scope} indicates one's scope of work, such as \textit{enterprise, project} and \textit{national}; lastly, \textit{Content} indicates one's content of work, such as \textit{data, r\&d} and \textit{security}. 

Finally, \textbf{Location} indicates the geographic scope that the title owner is responsible of. Examples of this NE tag include geographic regions such as \textit{APAC, Asia, European}, and counties/states/cities such as \textit{China, America} and \textit{Colorado}.
 
Formally, we define the occupational domain-specific NE tags as \textit{RES, FUN, LOC} and \textit{O}, indicating the \textit{responsibility, function, location} and \textit{others} respectively. For instance, a job title of \textit{chief financial officer asia pacific} is tagged as \textit{S-RES S-FUN S-RES B-LOC E-LOC} with the BIOES scheme~\cite{Ratinov2009design}. The distribution of the four labels are shown in Table~\ref{tab:labels}. We adopt a knowledge-based NE tagging strategy by creating a gazetteer of word tokens. This is achieved by first running a Unigram analysis of the job titles, sorted in descending order. Subsequently, the top 1,500 tokens are tagged by three annotators, who are a HR personnel, a senior recruiter and a seasoned business professional. Among 1,500 tokens tagged, every tag is agreed with at least two annotators, where 1,169 (77.9\%) are commonly agreed among all three annotators, and 331 (22.1\%) are agreed with two annotators. We further assess the Inter-Rater Reliability with two inter-coder agreements, achieving 0.853 on Percentage Agreement~\cite{viera2005understanding} and 0.778 on Cohen’s Kappa~\cite{artstein2008inter}, a \textit{Strong} level of agreement. Finally, the job titles are labelled with NE tags using BIOES scheme and formatted for NER tasks.

\begin{table}[h]
\centering
\begin{tabular}{ll}
\hline
Responsibility & \begin{tabular}[c]{@{}l@{}}Managerial level: \textit{lead, supervisor, manager, director, president}\\ Operational role: \textit{engineer, designer, accountant, technician}\\ Seniority: \textit{junior, vice, associate, assistant, senior}\end{tabular} \\ \hline

Function & \begin{tabular}[c]{@{}l@{}}Departments: \textit{sales, marketing, finance, operations, strategy}\\ Scope: \textit{enterprise, project, customer, national, site}\\ Content: \textit{data, r\&d, security, training, integration, education}\end{tabular} \\ \hline

Location & \begin{tabular}[c]{@{}l@{}}Regions: \textit{APAC, SEA, Asia, European, north, central}\\ Countries/States/Cities: \textit{China, America, Singapore, Colorado}\end{tabular} \\ \hline
\end{tabular}%
\caption{Examples of occupational NE tags.}
\label{tab:NEsample}
\end{table}

\section{Applications of IPOD: Named Entity Recognition and Job Embedding}

To demonstrate the usefulness of IPOD, we use this dataset for an occupational NER task and propose a job title vector representation called Title2vec, which we describe next.

\subsection{NER Models}

We propose two encoders to address the occupational NER task. One of the models is Conditional Random Fields (CRF), which is a probabilistic machine learning model that produce joint probability of the co-occurrence of output sequence~\cite{lafferty2001conditional}. The other model falls in the recurrent neural networks family, namely a bidirectional Long Short-Term Memory model with a CRF layer adding to the output layer (LSTM-CRF). Variations of the LSTM-CRF model showed state-of-the-art results in some recent works for different downstream NLP tasks~\cite{lample2016neural,liu2018empower}. Both the CRF and LSTM-CRF are decoded using a first-order Viterbi algorithm~\cite{forney1973viterbi} that finds the sequence of NE tags with highest scores.
We also construct two baseline encoders, namely a Logistic Regression (LogReg) classifier and a standard LSTM, both of which are decoded using a softmax layer.

\begin{figure*}[t]
  \includegraphics[width=\textwidth]{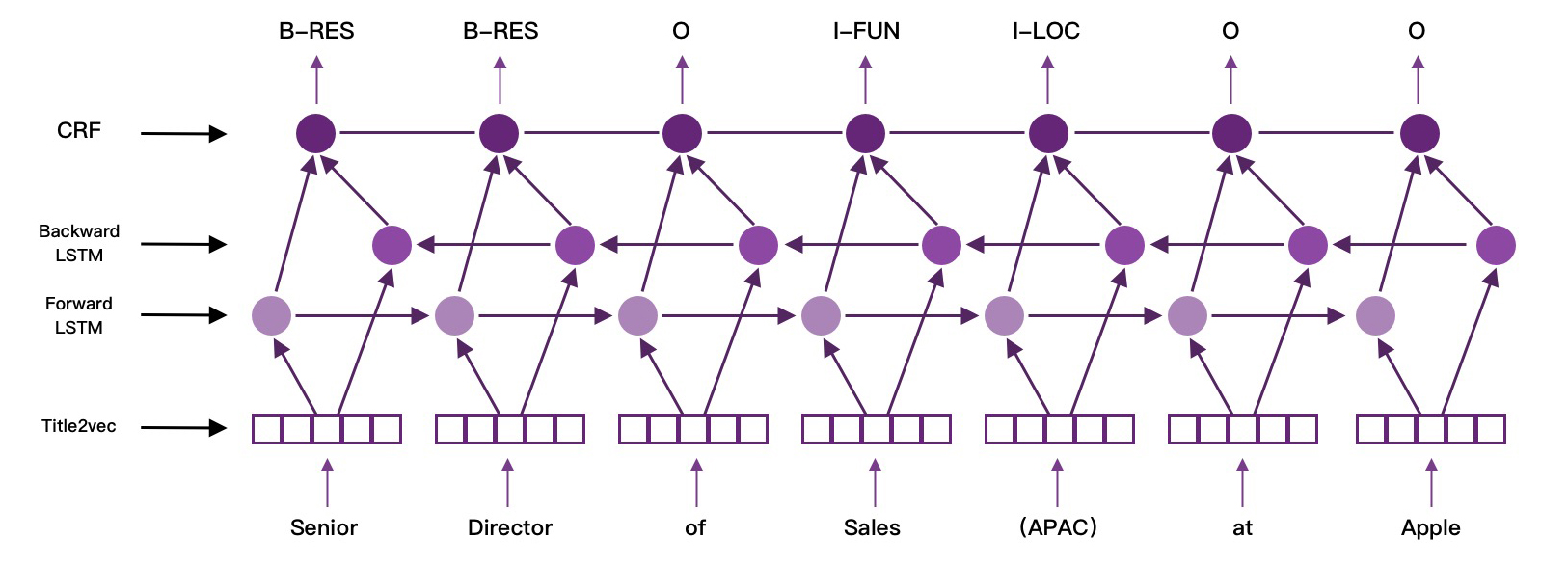}
  \centering
  \caption{Bidirectional LSTM-CRF model for occupational NER task}
  \label{fig:model}
\end{figure*}

\subsection{Job Embedding}

We propose a contextual job title vector representation model, \textit{Title2vec}, using a bidirectional Language Model (biLM) approach~\cite{peters2018deep} where each token is represented with a contextual vector with 3072 dimensions, before passing through a bi-directional LSTM network. The forward LSTM predicts the probability of each token given its history, and the backward LSTM takes the same approach with reverse order. 

Instead of training from scratch, we construct \textit{Title2vec} by fine-tuning from a pre-trained model, namely the Embedding from Language Models (ELMo)~\cite{peters2018deep}. The choice is because ELMo provides a language-level contextual meaning for word tokens that is highly similar, if not identical, to that in job titles. For instance, the word \textit{director} appearing in a job title is the same as that appears in a Wikipedia article. 


\subsection{Hyper-parameter Optimization}

We conduct grid search of hyper-parameters for both CRF and LSTM-CRF models to fine-tune the performance of proposed methods. The search space includes varying learning rates (i.e.,  $0.1$ or $0.01$), number of LSTM hidden layers (1 or 2), number of hidden states (128 or 256), mini-batch size (32 or 128) and type of optimizer(Adam or SGD). Word Dropout and Variational Dropout~\cite{kingma2015variational} are used to prevent over-fitting, with probability of 0.05 and 0.5 respectively. In total, we evaluate over 100 hyper-parameters sets, where each set of hyper-parameters is run with 10 epochs. 

\begin{table}[h]
\centering
\begin{tabular}{ccccc}
\hline
\multirow{2}{*}{Hyper Parameter} & \multicolumn{2}{c}{CRF} & \multicolumn{2}{c}{LSTM-CRF} \\  
 & Final & Range & Final & Range \\ \hline
Learning Rate & 0.1 & \{0.01,0.1\} & 0.1 & \{0.01,0.1\} \\ 
Mini-batch Size & 32 & \{32,128\} & 128 & \{32,128\} \\ 
Word Dropout & 0.05 & - & 0.05 & - \\ 
Variational Dropout & 0.5 & - & 0.5 & - \\ 
Type of Optimizer & SGD & \{Adam, SGD\} & SGD & \{Adam, SGD\} \\ 
LSTM Layers & - & - & 1 & \{1,2\} \\ 
LSTM State Size & - & - & 256 & \{128,256\} \\ \hline
\end{tabular}
\caption{Hyper-parameter search space and final values used}
\label{tab:hyperpara}
\end{table}

Table~\ref{tab:hyperpara} shows the breakdown of the search space and the final hyper-parameters used for both models. We deploy a Cross Entropy loss function and a SGD optimizer with an initial learning rate of $0.1$ and a mini-batch size of $32$ for both proposed CRF model and baselines. For the two LSTM-based models, we use a single hidden layer with an initial learning rate of $0.1$, LSTM state size of 256 and a mini-batch size of $128$.

\section{Experiments and Results}
\label{sec:experiment}

In this section, we discuss our experimental evaluation and results.

\subsection{Metrics}

Our work uses two metrics to assess performance of various machine models and human performance, namely \textit{Exact Match (EM)} and the \textit{F1} score, formally defined as $F1 = 2 * Precision * Recall * (Precision + Recall)^{-1}$. The \textit{EM} metric measures the percentage agreement between the ground truth and predicted labels with exact matches, while the \textit{F1} score metric is designed to measure the average overlaps between the ground truth and prediction. Furthermore, the overall \textit{Precision} and \textit{Recall} metrics of all models are also reported.

\subsection{Human Performance}

We construct the human performance baseline for IPOD using the NE tags annotated by the three domain experts. We choose the set of labels tagged by annotator 1 as the ground truth labels and compute against the other two annotation sets. We then take the average \textit{EM} and \textit{F1} to indicate human performance. We record an \textit{EM} accuracy of 91.3\%, and an \textit{F1} of 95.4\%. This shows a strong human performance as compared to those of other datasets, such as 91\% \textit{EM} for the CHEMDNER corpus\cite{Martin2015The}, 86.8\% EM and 89.5\% F1 for SQuAD2.0~\cite{Rajpurkar2018know}, and 77.0\% \textit{EM} and 86.8 \textit{F1} for SQuAD1.0~\cite{RajpurkarZLL16}.

\subsection{Model Performance}

Table~\ref{tab:results} shows the overall performance of our models and human performance on IPOD, in terms of precision(P), recall(R), exact match(EM) and F1. While the performance of LSTM, CRF and LSTM-CRF models are very close to each others ($\pm{0.2}$ difference), all three models outperform human in precision, recall, exact match and F1. 

\begin{table}[h]
\centering
\begin{tabular}{c|cccc}
\hline
Models      & P     & R     & EM    & F1    \\ \hline
LogReg      & 90.80 & 93.20 & 85.10 & 92.00 \\
LSTM        & 99.71 &  99.90  & 99.61 &  99.80    \\
CRF         & \textbf{99.90} &  99.81  & 99.71 & 99.85 \\
LSTM-CRF    & 99.86  & \textbf{99.97}  & \textbf{99.83}  & \textbf{99.91} \\ \hline
Human       & 91.60  & 99.60 &  91.30 & 95.40  \\ \hline
\end{tabular}
\caption{Overall results of Job Title NER}
\label{tab:results}
\end{table}

Table~\ref{tab:bytags} shows the per-tag breakdown of NER results, in terms of EM and F1. CRF and LSTM-CRF perform similarly to each others, and outperform LogReg and LSTM for all three categories, though LSTM model also has a good performance. CRF also shows a significant advantage in classifying RES tags (99.99 EM and 99.99 F1).

\begin{table}[h]
\centering
\begin{tabular}{@{}c|cccccc@{}}
\toprule
\multirow{2}{*}{} & \multicolumn{2}{c}{FUN} & \multicolumn{2}{c}{LOC} & \multicolumn{2}{c}{RES} \\ \cmidrule(l){2-7} 
 & EM & F1 & EM & F1 & EM & F1 \\ \midrule
LogReg & 78.30 & 87.80 & 93.70 & 96.80 & 90.10 & 94.80 \\
LSTM & 99.49 & 99.74 & 97.68 & 98.83 & 99.77 & 99.88 \\
CRF & 99.35 & 99.67 & \textbf{98.96} & \textbf{99.48} & \textbf{99.99} & \textbf{99.99} \\
LSTM-CRF & \textbf{99.88} & \textbf{99.94} & 98.70 & 99.35 & 99.82 & 99.91 \\ \bottomrule
\end{tabular}
\caption{Performance stratified by NE tags (EM, F1)}
\label{tab:bytags}
\end{table}

\section{Related Work}
\label{sec:related-work}

In this section, we review related works, in the area of occupational data mining and analysis, contextual embedding and Named Entity Recognition (NER).

\subsection{Occupational Data Mining and Analysis}
\label{sec:camo}

Prior works on occupational data mining and analysis aim to accomplish a wide range of tasks, such as Career Modeling and Job Recommendation. In the area of Career Modeling, prior works address downstream tasks including career path modeling~\cite{liu2016fortune,mimno2008modeling}, career movement prediction~\cite{james2018prediction,yang2018one}, job title ranking~\cite{xu2018extracting}, and employability~\cite{massoni2009career}. In Job Recommendation, past works focus on analysing Person-Job Fit~\cite{zhu2018person,shen2018joint,qin2018enhancing} which commonly aims to suggest employment suitability for companies, and Job Recommendation~\cite{malinowski2006matching,zhang2014research} which on the other hand provides decision analysis for the job seekers. These works commonly leverage real-world data from different sources, including Linkedin~\cite{liu2016fortune,li2017nemo}, resumes~\cite{mimno2008modeling,yang2018one}, job portals~\cite{zhang2014research,zhu2018person} and tech companies~\cite{shen2018joint}. 

The proposed solutions to these problems are based on different approaches. Most works utilize various machine learning approaches, such as linear classification models~\cite{liu2016fortune,li2017prospecting,james2018prediction,yang2018one}, generative models~\cite{mimno2008modeling,xu2018extracting} and Neural Networks~\cite{mimno2008modeling,zhu2018person,qin2018enhancing}. Some take algorithmic approaches, such as statistical inference~\cite{mimno2008modeling,james2018prediction,shen2018joint}, Graph-theoretic models~\cite{massoni2009career,paparrizos2011machine} and recommmender systems with content-based and collaborative filtering~\cite{zhang2014research,malinowski2006matching}. Some works report their time complexity to be polynomial~\cite{liu2016fortune,li2017prospecting}.

\subsection{Natural Language Processing}

\textbf{Word Embedding.} 
Classic word embedding methods construct word-level vector representations to capture the contextual meaning of words. A Neural Network Language Model (NNLM) was proposed with Continuous Bag of Words (CBoW) model and skip-gram model~\cite{bengio2003neural}, which lead to a series of NNLM-based embedding works~\cite{turian2010word}. Pennington et al., 2014 proposed GloVe~\cite{pennington2014glove}, which uses a much simpler approach, i.e., constructing global vectors to represent contextual knowledge of the vocabulary, that achieves good results. More recently, a series of high quality contextual models are proposed, such as ELMo~\cite{peters2018deep}, FastText~\cite{bojanowski2017enriching} and Flair~\cite{akbik2018contextual}.  Both word-level contextualization and character-level features are commonly used for these works.

\textbf{Document Embedding.}
While Word Embedding constructs static continuous vectors on wordlevel, recent works also propose methods to represent document-level embeddings. A transformer-based approach receives high popularity in recent literature. It uses pre-trained transformer-based models with very large datasets to construct the document-level embeddings, such as Bert~\cite{devlin2018bert} and GPT~\cite{radford2019language}, among others. This approach enables contextual embedding in both word level and document level. Lample et al. (2016) proposes a remarkable Stacked Embedding approach that constructs a hierarchical embedding architecture of document-level embedding by stacking word-level embedding with character-level features and concatenating with an RNN, which performs well in NER tasks~\cite{lample2016neural}. 


\textbf{Named Entity Recognition.} 
Named entity recognition is a challenging task that traditionally involves a high degree of manually crafted features for different domains. The good news is that numerous large-scale corpora, such as CoNLL-2003~\cite{sang2003introduction} and Ontonotes~\cite{weischedel2013ontonotes},  are made available for training with deep neural architectures. State-of-the-art NER models are LSTM-based~\cite{lample2016neural,liu2018empower}, where feeding the sentence embeddings into uni- or bi-directional LSTM encoder. Instead of decoding directly, some works also add a Conditional Random Field (CRF) layer at the end while training the classifier, and use Viterbi~\cite{forney1973viterbi} to decode the probabilistic output into NE labels. The recently popular transformer-based models~\cite{devlin2018bert,radford2019language} are also capable of producing good results.

While manually tagging a large dataset requires tremendous amount of efforts, prior works leverage knowledge-based gazetteers developed by various unsupervised or semi-supervised learning approaches~\cite{kazama2008inducing,saha2008gazetteer}, or rely on generative models~\cite{nallapati2010blind,mukund2009ne}. Tags can be further formatted with tagging schemes such as IOB~\cite{ramshaw1999text} or BIOES~\cite{Ratinov2009design}, to indicate the position of tags in a chunk.

\section{Conclusion}
\label{sec:conclusion}

In this work, we present the IPOD corpus that comprises a large number of job titles, with a knowledge-based gazetteer that includes manual NE tags from three domain exports annotators. We also address two challenging upstream tasks of occupational data mining and analysis, namely job title embeddings and occupational NER. Despite strong human performance records of 91.3\% \textit{EM} and 95.4\% \textit{F1}, our proposed models, namely CRF and bidirectional LSTM-CRF, outperform human and baselines in EM and F1 for overall results and per-tag breakdown.  Finally, we release a pre-trained \textit{Title2vec} job title vector representation that can serve as basic building blocks and improve the performance for a wide spectrum of downstream tasks. To the best of our knowledge, our work is the first attempt to address the challenging occupational NER task and both the dataset and pre-trained embeddings are first made available in the literature of occupational analysis. 

%
%
%
\bibliographystyle{splncs04}
\bibliography{ref}
\end{document}